\documentclass[letterpaper,10pt,conference]{ieeeconf}  

\IEEEoverridecommandlockouts                              

\usepackage{amsfonts, algorithm} 
\usepackage{amsmath,amssymb}
\usepackage{algpseudocode}
\usepackage{graphicx}
\usepackage{wrapfig}
\usepackage{xspace}
\usepackage{xcolor}
\usepackage{booktabs}
\usepackage{pifont}
\usepackage{hyperref}
\DeclareMathOperator*{\argmin}{arg\,min}
\newcommand\norm[1]{\left\lVert#1\right\rVert}

\newcommand{\nn}{{\mathbf{NN}}}

\newcommand{\name}{{WayEx\xspace}}

\usepackage[style=numeric,natbib,hyperref=true,sorting=none]{biblatex} 
\title{\name: Waypoint Exploration using a Single Demonstration}

\author{Mara Levy, Nirat Saini, and Abhinav Shrivastava\\[0.5em]
University of Maryland, College Park
}

\begin{document}

\maketitle
\thispagestyle{empty}
\pagestyle{empty}

\begin{abstract}
We propose WayEx, a new method for learning complex goal-conditioned robotics tasks from a single demonstration. Our approach distinguishes itself from existing imitation learning methods by demanding fewer expert examples and eliminating the need for information about the actions taken during the demonstration. This is accomplished by introducing a new reward function and employing a knowledge expansion technique. We demonstrate the effectiveness of \name, our waypoint exploration strategy, across six diverse tasks, showcasing its applicability in various environments. Notably, our method significantly reduces training time by $\sim$50\% as compared to traditional reinforcement learning methods. WayEx  obtains a higher reward than existing imitation learning methods given only a single demonstration. Furthermore, we demonstrate its success in tackling complex environments where standard approaches fall short. Appendix is available at: \href{https://waypoint-ex.github.io}{https://waypoint-ex.github.io}.

\end{abstract}

\section{INTRODUCTION}
Humans have a natural ability to learn tasks by observing a single demonstration which they can follow step-by-step. For instance, we can watch a video and grasp how to open a vault, then practice until we succeed without requiring further instructions. Drawing inspiration from this ability, the combination of learning from demonstrations and reinforcement learning techniques has become a popular and potent approach for training robots~\cite{OvercomingExploration,Montezuma,Visuomotor,DemoRL}. However, compared to humans who can learn simple tasks from a single demonstration, robots require a multitude of diverse expert instances. For example, to learn how to open a vault, the robot must observe successful demonstrations for different views and locations of the vault handle with respect to the robot's location. Moreover, each demonstration must contain information about the location of the vault (state), the precise joint rotations (action) to reach the vault, and knowledge about how close it is to completing the task (reward).
Hence, most common methods in learning from demonstrations, such as Imitation learning~\cite{DeepMimic, waypoints, DemoRL} and Inverse reinforcement learning~\cite{GAIL,SQIL,airl} require a set of expert demonstrations, with a defined state, action and reward space. Collecting all the data in real time and computing the definitive action and reward space is impractical and inefficient. Therefore, in this work, we strive to reduce these inefficiencies by using only a single demonstration for training. Additionally, our setup does not require knowledge about the action space, relying on just the state space and a corresponding reward function.

\begin{figure}[htb]
\centering
\includegraphics[width=\linewidth]{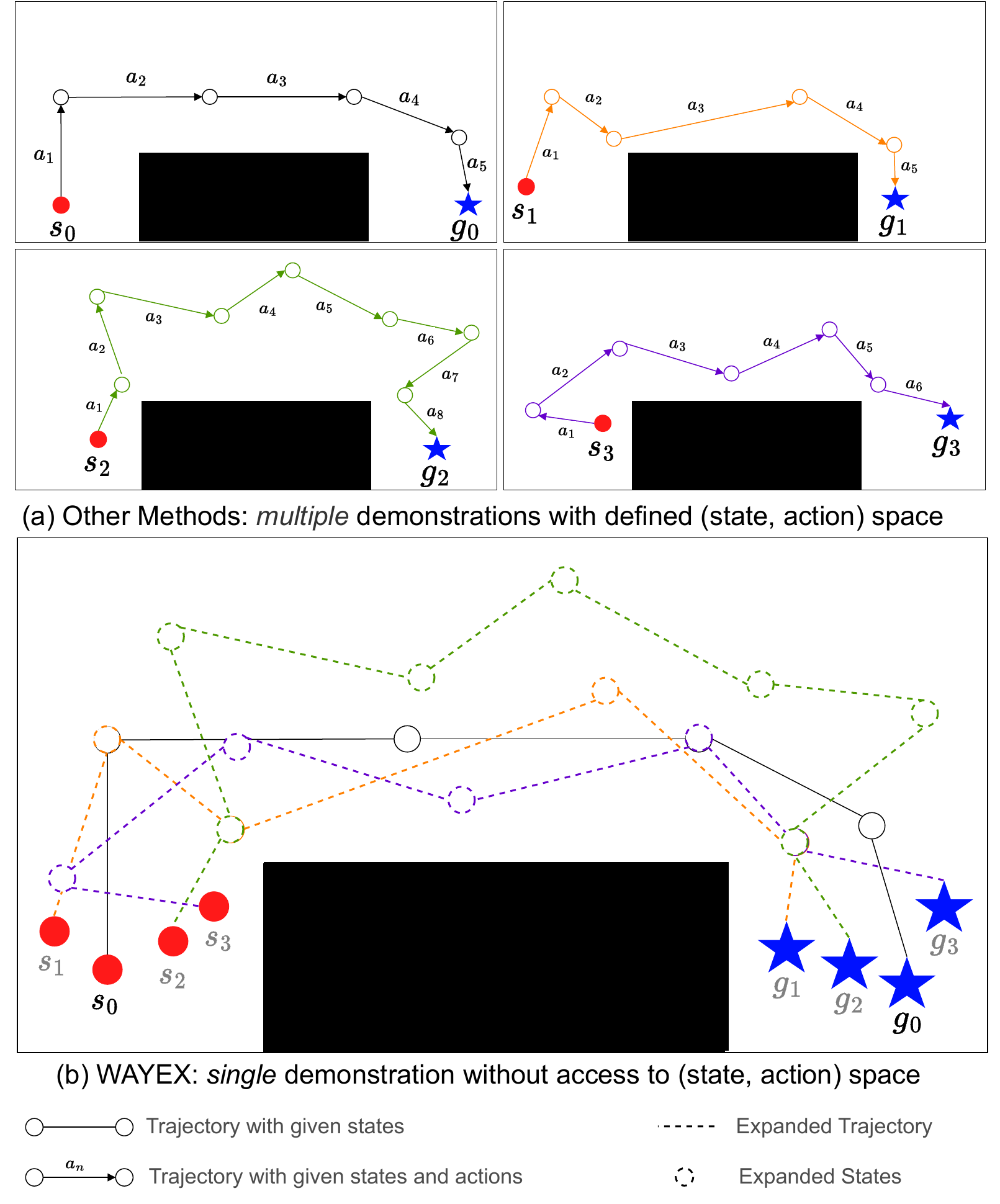}
\vspace{-1.5em}
\caption{A comparison of our approach to general imitation learning techniques. \textbf{(a)} Traditional Imitation learning approaches require multiple expert trajectories with a known action space for training (4 shown here). \textbf{(b)} For our proposed method \name\, we use only one expert trajectory, and expand knowledge from this one trajectory to learn how to solve the task. During training with a single initial state ($s_0$) and a single goal state ($g_0$), our model learns to navigate back to the expert trajectory from points that are not part of the trajectory (all dotted states, which can be a combination of 4 expert trajectories shown on the top). This enables the model to successfully reach the goal state. We further introduce additional start and goal states ([$s_1, g_1$],[ $s_2,g_2$], [$s_3,g_3$]).}
\label{fig:teaser}
\vspace{-2em}
\end{figure}
Prior works identify key states (waypoints~\cite{waypoints, DeepMimic}) along the robot's trajectory, to help it navigate towards the goal. We also employ waypoints to solve the task, without having access to the action space. Each observation within the demonstration is defined as an individual waypoint. While other approaches~\cite{waypoints} need access to waypoints during the testing phase, \name\ only requires waypoints during training. This prevents the need for training a model to predict the waypoints during inference. We leverage the waypoints during training via an augmented reward structure based on the known Q-Values~\cite{q-learning} associated with each waypoint. 

A common approach to solving a task without a dataset of expert trajectories is to use dense rewards, based on the key steps of a task. Existing studies in the field of reinforcement learning acknowledge that employing dense rewards is difficult since it requires practitioners to engineer specific reward functions for each task. Additionally these rewards, if ill-designed, can lead to unforeseen behavior. To circumvent these challenges, we assume a sparse reward structure when determining the new reward. With sparse rewards, the robot receives a reward of 0 if the actions lead to a goal state; otherwise, the reward is -1. Acquiring this reward solely through the process of exploration can prove to be highly challenging, making certain tasks unachievable with sparse rewards alone~\cite{atari}. Our approach strikes a balance, allowing the model to receive frequent rewards without exposing it to the typical risks associated with dense rewards.

By utilizing this new reward structure, our model can solve tasks that closely resemble the demonstration setup. However, despite its ability to learn from a single example, our approach still encounters a common limitation of learning from demonstrations. If the robot encounters a state beyond the scope of what it has previously encountered it will not know how to proceed. To overcome this challenge, the robot needs to acquire experience beyond the confines of the provided demonstrated space. A commonly employed method to achieve this is to integrate learning from demonstrations with conventional model-free reinforcement learning algorithms~\cite{DDPG,SAC}. Strict model-free reinforcement learning involves learning optimal state-action pairs through trial and error rather than relying on expert trajectories. We combine model-free reinforcement learning with our waypoint reward and introduce an additional expansion method that further enhances the model's knowledge.

In this work, we propose \name, derived from \textbf{Way}point \textbf{Ex}ploration, a novel approach that enables the training of a reinforcement learning model using a single expert demonstration and without any prior knowledge of the action space. It can serve as a wrapper around any reinforcement learning algorithm, facilitating its applicability as the field advances. Our primary technical contributions are: (1) the introduction of a new reward function based on sparse rewards, which provides additional rewards without introducing unforeseen consequences, and (2) a method for expanding knowledge beyond a single demonstration to encompass the entire spectrum of both the state and goal spaces. We demonstrate that our approach enables faster learning of tasks compared to previous reinforcement learning methods while requiring minimal additional information.

\section{RELATED WORK}
\label{sec:related_work}
\textbf{Goal-Conditioned Reinforcement Learning}. We are interested in investigating tasks that involve a robot reaching a specific end state specified by an initial ``goal.'' In the reinforcement learning (RL) community, these problems are known as Goal-Conditioned Tasks. Prior works have studied how to use RL in many different ways in order to solve these tasks~\cite{HER, Fetch, Planning, GCRL, Sutton, CHER}. Early works such as~\cite{Fetch, Sutton} show that it is possible to use standard RL methods such as~\cite{DDPG, SAC, qlearning}, but it can be time-consuming, and there are some types of tasks that these methods alone cannot solve. To combat this, other works have suggested the use of hindsight re-labeling~\cite{HER, actionable}, which speeds up the process, but still requires a large amount of data to reach a successful trajectory~\cite{Fetch}.

\begin{figure}[t!]
\centering
\includegraphics[width=1.0\linewidth]{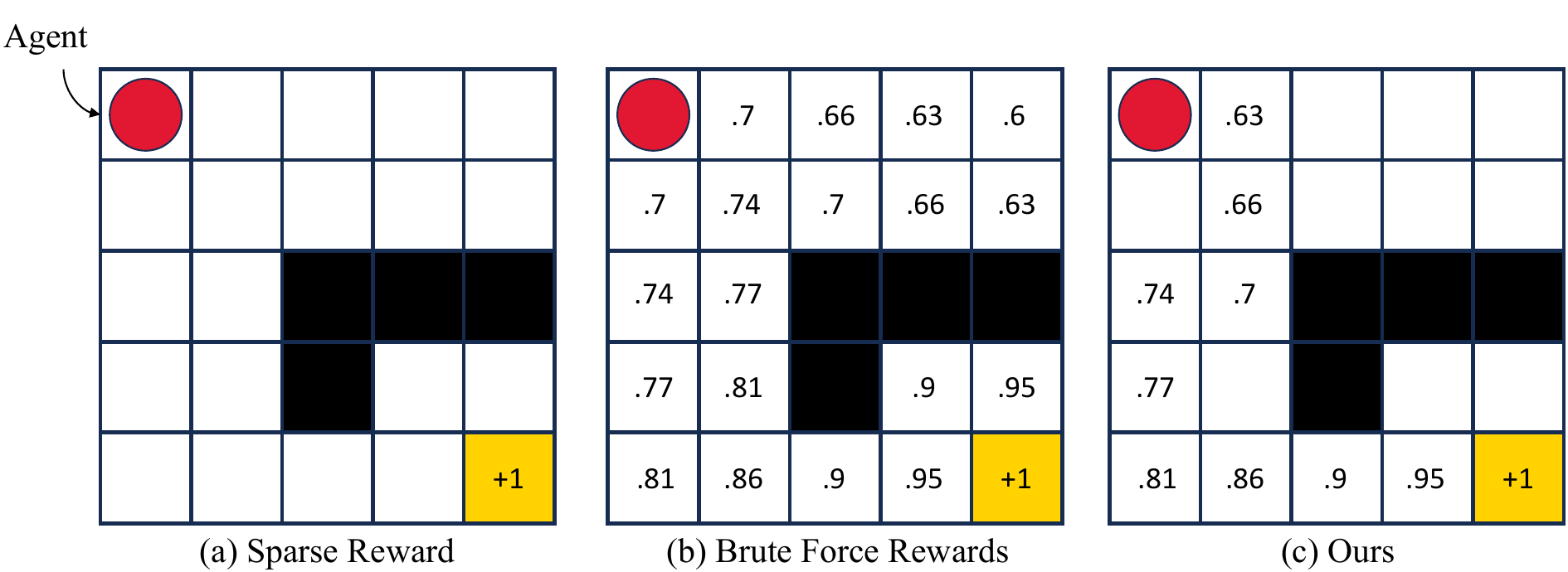}
\caption{\textbf{Visualization of Grid World Toy Example.} \textbf{(a)} shows the environment setup with a sparse reward. \textbf{(b)} shows the reward for each state once the entire environment has been solved using the bellman equation~\cite{bellman}. \textbf{(c)} represents \name\ where with a single demonstration, we compute a close approximation of the reward for each state along the path to the goal.}
\label{fig:method}
\end{figure}

\textbf{Imitation Learning.} Learning from demonstrations, also known as imitation learning, is a common approach for solving goal conditioned tasks. Our approach uses ideas similar to prior works such as~\cite{DeepMimic, DemoRL, Montezuma, Visuomotor, OvercomingExploration, HardExplore}. These works take recorded successful episodes of a task and use them to aid in training the RL model. Several of these papers operate by adding successful demonstrations to the replay buffer~\cite{OvercomingExploration, HardExplore}; however, these papers require accurate knowledge of the action space. This type of information is only accessible in datasets specifically designed for robot training. In order to learn from other sources, such as videos, the field must evolve past this method. Additionally, aside from~\cite{DeepMimic, Montezuma}, these works all require a large number of demonstrations. ~\cite{DeepMimic} can use a smaller number, but needs a task-specific reward function.~\cite{Montezuma} only uses one example, but requires different start states.~\cite{DemoRL} proposes pre-training the model and then fine tuning for each individual task, but their pre-training is data intensive and not always generalizable.

\textbf{Inverse Reinforcement Learning.} Another closely related area, Inverse Reinforcement Learning (IRL), uses a model to predict the reward values based on the state and action space. Methods such as~\cite{GAIL, airl, SQIL}, employ a hybrid approach involving both IRL and adversarial learning to solve a task with limited demonstrations. Similar to our method, IRL methods are useful because they do not require a defined reward function. Nonetheless, despite their claim of using a small number of demonstrations, these methods still require at least 50 demonstrations. Additionally, these methods struggle to generalize beyond the initial demonstrations. Our approach differs from IRL approaches because we do not require the training of a model to define our generalized reward, which leaves less room for error and requires less data. 

\textbf{Modified Sparse Reward.} Several methods~\cite{MCAC, SACX, KeepingDistance, Lu2019PredictiveCF} attempt to modify the sparse reward function in different ways to make learning more efficient. Despite employing a similar reward structure to ours and looking at the maximum of two different functions, ~\cite{MCAC} still requires a large number of demonstrations as well as access to the action space. Another way to learn with sparse rewards is to split multi-step tasks into several different tasks~\cite{SACX}. This allows the sparse rewards to be more frequent, but requires a precise definition of each auxiliary task which makes creating a generalized model more difficult. 
\section{Method}
\subsection{Preliminaries}
We formulate our problem as a Markov Decision Process (MDP) consisting of an [n]-tuple $(S, A, R, \tau_D, \gamma)$. The elements of this tuple are the state space $S$, the action space $A$, the reward function $R\colon S \times A \rightarrow \mathbb{R}$, the demonstration trajectory $\tau_D\colon (s^*_0, \dots, s^*_N)$ and the discount factor $\gamma$. We will refer to a random episode trajectory as $\tau$, where $\tau = (s_0, a_0, \dots, s_N, a_N)$. $N$ is the total number of states and actions to reach every state. A policy is represented as $\pi_\theta\colon S \rightarrow A$, with parameters $\theta$. We use a sparse reward paradigm for our method, where the action space is unknown. A sparse reward is defined as a reward function $R$ that receives a reward of $0$ when in the goal state, $g$. At all other times the reward is $-1$. If $g$ represents the goal state, and $s_n$ and $a_n$ are the $n^\text{th}$ state and action respectively, then the sparse reward function $R$ can be represented as
\begin{equation}
\label{eq:sparse}
R(s_n, a_n)=
    \begin{cases}
        0 & \text{if} \; s_n = g,\\
        -1 & \text{if} \; s_n \neq g.
    \end{cases}
\end{equation}

\subsection{Overview}
In this section, we describe our method \name\ for learning goal-conditioned skills from a single demonstration with no information about the demonstration's action space. An illustrative overview of our method is provided in Figure~\ref{fig:method}, which shows a much simpler grid world demonstration. The leftmost grid in Figure~\ref{fig:method} represents the environment, where an agent must traverse the boxes to find the goal, which gives a sparse reward of 0. The middle grid shows the reward for each box using a bellman equation~\cite{bellman}, which requires access to rewards for all states. Finally, the right grid shows our approach which traverses a single path and then recursively determines the reward of each state along the path. Access to one successful demonstration allows \name\ to determine the pseudo ground truth rewards for each box along the path.



\name\ uses a sparse reward, along with bellman's equation to compute the value for each waypoint along the demonstration path. During exploration, a new state obtains a reward if its distance from a known waypoint is less than a threshold $d_\text{thresh}$ (captured by \texttt{is$\_$prox$\_$wp$()$}). We use Nearest Neighbors to determine the waypoint the new state is compared to. After training and achieving some success, we improve our method's ability to generalize to unseen start and goal states, by expanding on possible start and goal states. We present an algorithmic overview of \name\ in Algorithm~\ref{alg:reward}, followed by a comprehensive breakdown of each step for a clearer understanding.


\subsection{Proximal Waypoint}
Each state of the environment can be represented as $s_i = \{p_1, p_2, \dots, p_K\}$, $\forall$ i $\in (1,\dots,N)$, where $p_k$ represents an environmental parameter such as object pose and gripper velocity and $K$ represents the number of environmental parameters. Note that each parameter, $p_k$, is a relative value, with respect to the object's location, instead of an absolute value with respect to the world coordinate system. This ensures better generalizability of our method. We use our policy $\pi_{\theta}$ to determine an action that allows us to reach an unseen state $s_e$. Following other reinforcement learning algorithms~\cite{SAC, DDPG}, we add random noise to the action space, in order to increase the amount of exploration done during training. Once we have reached $s_e$ we will determine if it is within close proximity of a waypoint along our demonstration trajectory $\tau_D$. To do this we first use the Nearest Neighbor function to find the waypoint with the smallest total L2 distance between the parameters in the waypoint and the parameters in $s_e$ as
\begin{align}
s^*_t = \nn(s_e, \tau_D) &= \argmin_{\forall s^*_i \in \tau_D} \norm{s_e - s^*_i}_2,
\label{eq:nn}
\end{align}
where $s^*_t \in \tau_D$ is the closest state to (or the proximal waypoint for) $s_e$ from the states within the expert trajectory $\tau_D$. For an agent to receive a reward at state $s_e$, the distance between the parameters of $s_e$ and $s^*_t$ should be less than a threshold. For instance, if $d_\text{thresh} = \{d_1, d_2, \dots, d_K\}$ is the corresponding threshold for parameters between $s_e = \{p_1, p_2, \dots, p_K\}$ and $s^*_t = \{p^*_1, p^*_2, \dots, p^*_K\}$, then we compute the Boolean \texttt{hasProxWP} as
\label{eq:thresh}
\begin{align*}
\text{\texttt{is$\_$prox$\_$wp}}(s_e,s^*_t) =
    \begin{cases}
        \text{True},  & \text{if} \; \norm{p^*_i - p_i} \leq d_i, \forall i \in K,\\
        \text{False}, & \text{otherwise.}
    \end{cases}
    \end{align*}
We define one $d_\text{thresh}$ for each of the waypoints in $\tau_D$. For instance $s^*_t$ has its own threshold, $d^*_t$, and when $s^*_t$ is the nearest neighbor to $s_e$, we use $d^*_t$ as the threshold. In order to encourage progression, if \texttt{hasProxWP} is False 10 consecutive times for a waypoint $s^*_t$, then we increase $d^*_t$ by $\epsilon \;(= 0.001)$. We repeat this until we explore a point that falls within the threshold of the waypoint $s^*_t$.

\begin{algorithm}
    \caption{\name, prior to expanding knowledge}
    \label{alg:reward}
    \begin{algorithmic}[1]
        \State $\tau_D = (s^*_0, \dots, s^*_T):$ A successful demonstration
        \State $\pi_\theta\colon$ The policy that we will follow and update 
        \While{true}
            \State $\tau \gets (s_0, a_0, \dots, s_N, a_N)$ : an episode roll out
            \ForAll{$s_n, a_n \in \tau$}
                \State $s^*_t \gets \nn(s_n, \tau_D)$, where $s^*_t \in \tau_D$
                \State\texttt{hasProxWP} $ \gets $ \texttt{is$\_$prox$\_$wp}$(s_e,s^*_t)$
                \State $r \gets \text{reward}(\text{\texttt{hasProxWP}},\; t,\; l_\text{D}, l_\text{max})$ 
                \State $R \gets \max(r,\; \gamma * \text{critic}(s_{n + 1}))$ 
            \EndFor
        \EndWhile
    \end{algorithmic}
\end{algorithm}

\begin{figure*}[ht] 
\vspace{.2in}
\centering
\includegraphics[width=.9\linewidth]{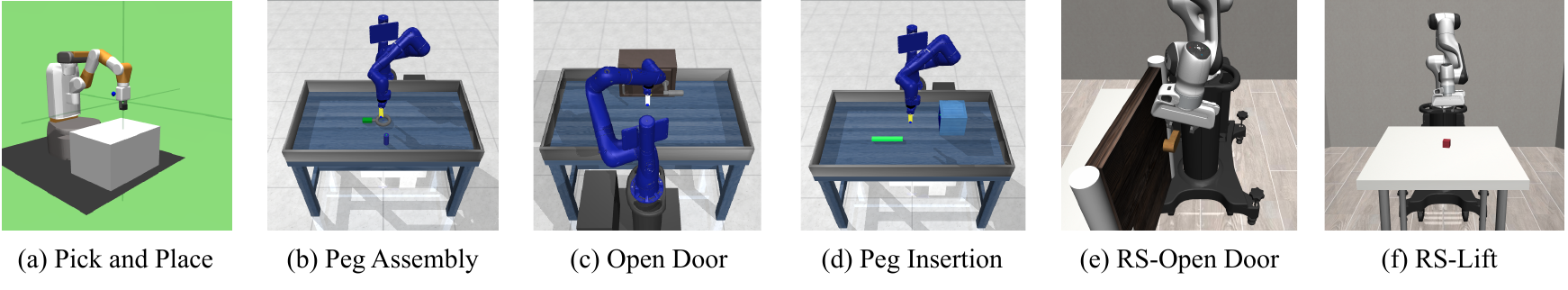}
\caption{The environments that we experimented on with \name. We show results on 4 different tasks: (a) pick and place, (b) peg assembly, (c) open door and (d) peg insertion. These tasks are ideal because they have a clear definition of success and therefore a clear sparse reward. However, most of these tasks cannot be solved with sparse rewards alone.}
\vspace{-0.15in}
\label{fig:envs}
\end{figure*}

\begin{figure*}[ht]
\centering
\includegraphics[width=.95\linewidth]{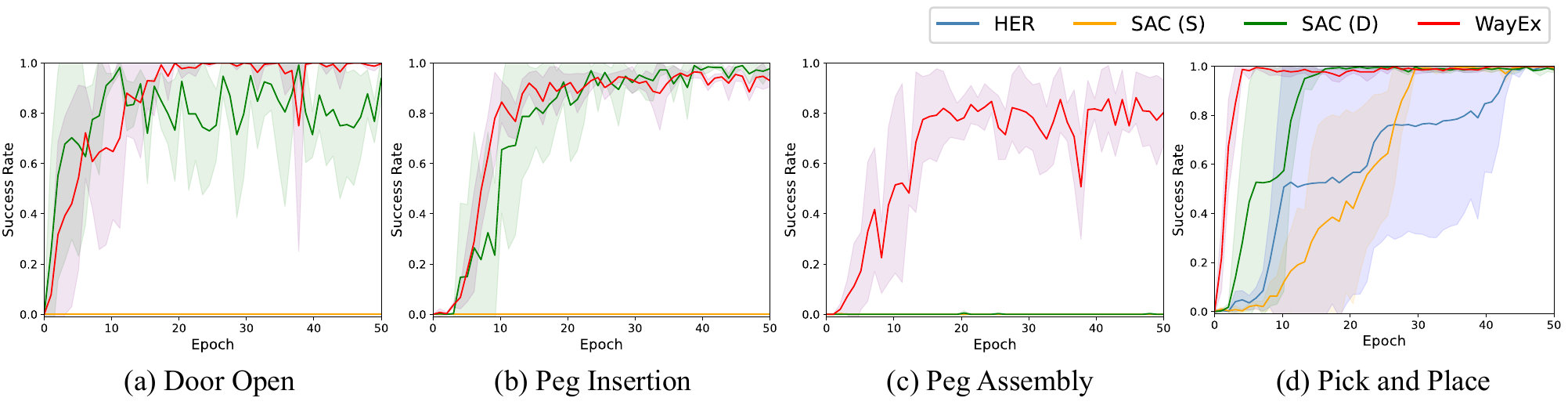}
\caption{\textbf{(a,b,c)} shows the results of the Meta World~\cite{metaworld} environments when trained using SAC~\cite{SAC} and a batch size of 2048. \textbf{(a)} Open Door Task, \textbf{(b)} Peg Insertion Task, \textbf{(c)} Peg Assembly Task. \textbf{(d)} Pick and Place task is from OpenAI~\cite{Fetch}.}
\label{fig:metaworldsac}
\end{figure*}

\begin{table*}[t]
\centering
\footnotesize
\caption{Requirements for different baselines and \name\ in terms of state, action and number of expert demonstrations.}
\begin{tabular} {@{}l c c c c c c c c c @{}}
\toprule
\multicolumn{1}{c}{Pre-requisites} & \multicolumn{1}{c}{SAC (S)} & \multicolumn{1}{c}{SAC (D)} & \multicolumn{1}{c}{HER} & \multicolumn{1}{c}{SAC + RB} 
& \multicolumn{1}{c}{SAC + MCAC} & \multicolumn{1}{c}{AWAC}
& \multicolumn{1}{c}{AWAC + MCAC} & \multicolumn{1}{c}{\textbf{\name}}\\
\midrule
Requires Action Space  & \ding{55} & \ding{55} & \ding{55} & \ding{51} & \ding{51} & \ding{51} & \ding{51} & \textbf{\ding{55}}\\
Requires Pre-training &  \ding{55} & \ding{55} & \ding{55} & \ding{55} & \ding{55} & \ding{51} & \ding{51} & \textbf{\ding{55}} \\
\# Expert Demonstrations  & 0 & 0 & 0 & 1 or 100 & 1 or 100 & 1 or 100 & 1 or 100 & \textbf{1}\\
\bottomrule
\end{tabular}
\label{table:methods}
\vspace{-0.2in}
\end{table*}
\subsection{New Reward Function}


Given the nearest neighbor waypoint, $s^*_t$, and the boolean result, \texttt{hasProxWP}, we can solve for the reward function, $r$, for our state action pair $(s_e, a_e)$.  $l_\text{max}$ represents the maximum length of an episode, $l_\text{D}$ represents the length of the demonstration $\tau_D$ and $t$ represents the time at which $s^*_t$ occurs. We compute the proposed reward, $r$ as
\begin{equation}
\label{eq:reward}
r = \begin{cases}
\displaystyle\sum_{i=0}^{l_\text{D} - t} -\gamma^i & \text{\texttt{hasProxWP},} \\
\displaystyle\sum_{i=1}^{l_\text{max}} -\gamma^i & \text{not \texttt{hasProxWP}.}
    \end{cases}
\end{equation}
If \texttt{hasProxWP} is false, we still want to account for the possibility that our state, $s_e$, is on a successful trajectory.
To do this, we say that the final reward $R = \text{max}(r, \gamma * \text{critic}(s_{e + 1}))$. In order for this to work we need to warm up the critic. We do this by training it for $1000$ time steps on just $r$ before we include the critic reward. For more information on actor critic methods please refer to~\cite{SAC, DDPG}.
\subsection{Expanding Knowledge}
Following the demonstration, \name\ teaches the policy how to solve the task from a fixed start and goal state. We now need to expand to every possible start and goal location. To achieve this, we slowly increase the number of possible start locations and goal locations, by adding random noise to the initial state space.

Given our current demonstration trajectory $\tau_D$ with states $s^*_0$ and goal $g^*$, let $\mathcal{N}^*(\mu^*,\,\sigma^*)$ be the distribution representing the amount of noise we will add to the start state and goal state. We will set the mean, $\mu^*$, to $0$ and only increase the standard deviation $\sigma^*$.

At the start of training $\sigma^*$ is set to 0.  $\sigma^*$ uses a modification strategy applied every 25 episodes. The updated value of $\sigma^*$ is described as $\sigma'$, where:
\begin{equation}
\label{eq:sg}
\sigma'=
    \begin{cases}
        \sigma^* + 0.001,  &   \text{if success rate} \geq 0.05,\\
        \sigma^*, & \text{otherwise.}\\
        
    \end{cases}
\end{equation}
\label{sec:method}
\begin{figure*}
\centering
\includegraphics[width=0.85\linewidth]{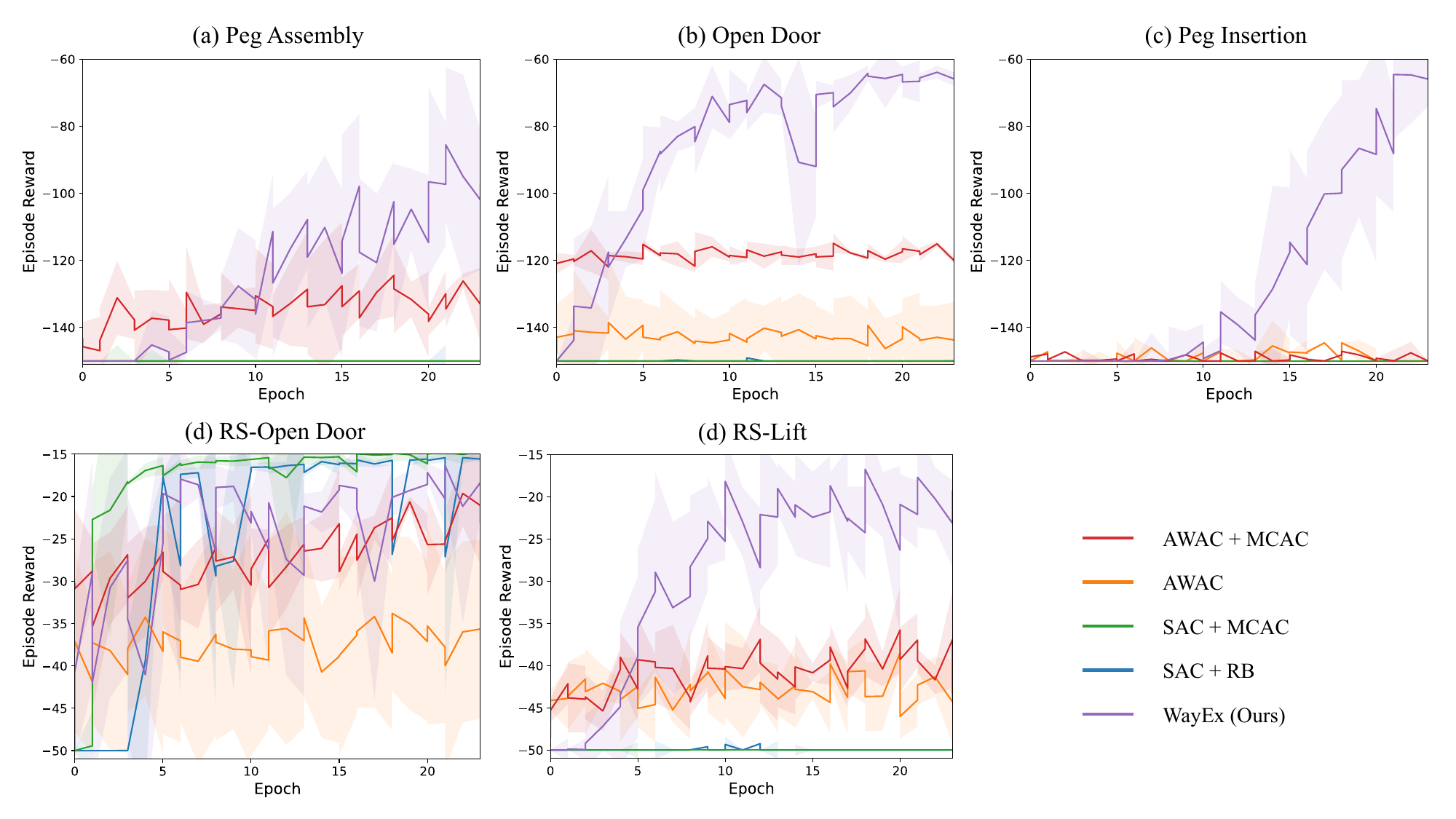}
\caption{This figure shows the results of several baselines when the are given one expert demonstration \textbf{(a,b,c)} shows the results of the Meta World~\cite{metaworld} environments when trained using AWAC~\cite{nair2021awac}, MCAC~\cite{MCAC}, SAC + RB and AWAC + MCAC. \textbf{(d,e)} shows the results of these same baselines on the robosuite tasks~\cite{robosuite2020} }
\label{fig:all1}
\vspace{-0.2in}
\end{figure*}
\section{Experiments and Results}

\subsection{Environment Setup}
\label{exp:setup}
We use MuJoCo~\cite{mujoco} to simulate our tasks. The implementation of our reinforcement learning algorithms was based on modified versions of the open-source code provided by stable-baselines3~\cite{stable-baselines3} and our baselines were based on ~\cite{MCAC}. Although we used only one demonstration for each task, we varied the starting demonstration across different seeds to demonstrate the adaptability of our method to different initial demonstrations. To ensure robustness, we conducted each experiment four times with different seeds and present the mean and standard deviation of these seeds in our graphs. Each epoch consists of 40,000 simulated timesteps in MuJoCo. The pick and place environment, the robosuite door environment and the robosuite lift environment have 50 time steps per episode, and the remaining environments have 150 time steps per episode. We pre-train AWAC for 25000 timesteps.
\subsection{Tasks}
\label{exp:experiments}
\name\ is trained on six distinct tasks, each designed to showcase the robot's capability to accomplish simple goal-conditioned objectives. The pick and place environment is from the OpenAI Fetch tasks~\cite{Fetch}, the next three (open door, peg insertion, and peg assembly) tasks are from Meta World~\cite{metaworld}. The final two tasks, Robosuite (RS)-Open Door and RS-Lift come from~\cite{robosuite2020}. Images of these tasks can be seen in Figure~\ref{fig:envs}. Our tasks are: \textbf{(a) Pick and Place:} The task is to grasp a box and move it towards a goal in the air. \textbf{(b) Peg Assembly:} The task is to pick up a round nut and then place the round part over a peg. \textbf{(c) Open Door:} The task is to grasp a door handle and then open the door until it reaches a goal location. \textbf{(d) Peg Insertion:} The task is to pick up a peg and insert it into a hole. \textbf{(e) RS-Open Door:} The task is to grasp a door handle and then open the door very slightly. \textbf{(f) RS-Lift:} The task is to pick up a block and lift it into the air.\looseness=-1

\subsection{Baslines}
We test our method (Table~\ref{table:methods}) against following baselines:
\begin{itemize}
    \item \textbf{SAC}. This baseline uses the Soft Actor Critic (SAC) algorithm~\cite{SAC} as the reinforcement learning algorithm. It can be initialized in three ways: (1) Sparse Reward \textbf{(SAC (S))}, (2) Dense Reward \textbf{(SAC (D))}, (3) SAC + Replay Buffer \textbf{(SAC + RB)}: we initialize the replay buffer with expert demonstrations.
    \item \textbf{Hindsight Experience Replay (HER)} is a well known technique~\cite{HER}, which uses the previous experiences to learn overtime.
    \item \textbf{SAC+MCAC}. This baseline uses the soft actor critic (SAC) algorithm~\cite{SAC} as well as Monte Carlo augmented Actor-Critic~\cite{MCAC}. 
    \item\textbf{Advantage Weighted Actor Critic (AWAC)} follows the methods described in~\cite{nair2021awac}. \item\textbf{AWAC+MCAC} uses~\cite{nair2021awac} along with a modified reward as describe in~\cite{MCAC}.
\end{itemize}

\subsection{Results}
\begin{figure*}
\centering
\includegraphics[width=0.85\linewidth]{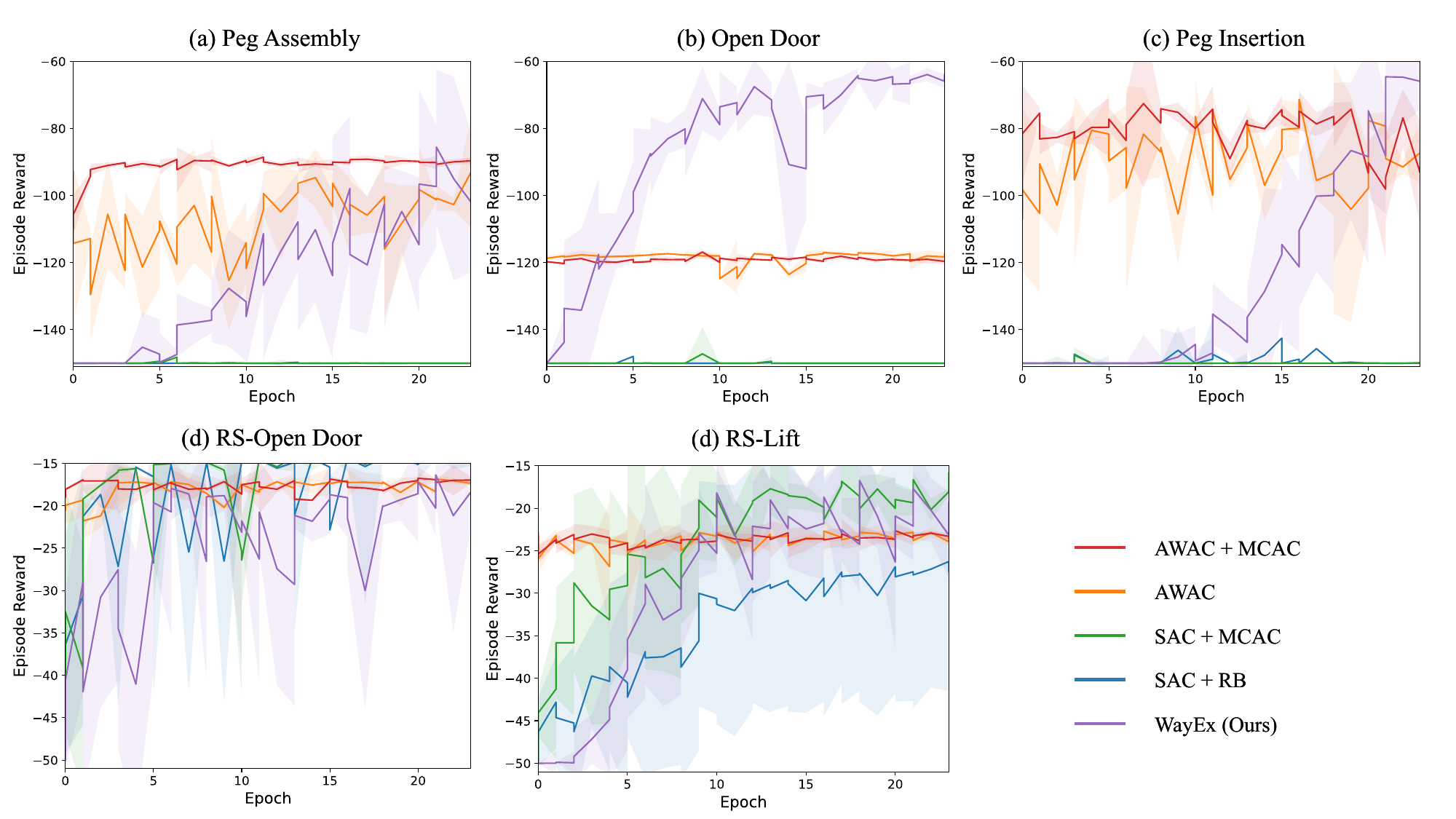}
\vspace{-0.1in}
\caption{This figure shows the results of several baselines when they are given one hundred expert demonstrations \textbf{(a,b,c)} shows the results of the Meta World~\cite{metaworld} environments when trained using AWAC~\cite{nair2021awac}, MCAC~\cite{MCAC}, SAC + RB and AWAC + MCAC. \textbf{(d,e)} shows the results of these same baselines on the robosuite tasks~\cite{robosuite2020}. Our method continues to use only one demonstration. }
\label{fig:all2}
\vspace{-0.2in}
\end{figure*}

We evaluate several different combinations of baselines each of which have different requirements in comparison to our method.
\subsubsection{No Action Space}
First we look at methods that do not require access to the action space of an expert demonstration in order to learn. In Figure~\ref{fig:metaworldsac}(d), we analyze the performance of our method compared to other conventional reinforcement learning algorithms for the pick and place task. These graphs reveal two significant observations: (1) \name\ exhibits a remarkable acceleration in the learning process with just a single demonstration, over SAC. (2) \name\ demonstrates a considerably lower standard deviation compared to the other algorithms. We hypothesize that in general, methods relying on sparse rewards requires a degree of luck. For that, the environment must be explored extensively until a reward is obtained, allowing the method to learn the task. However, \name\ circumvents this by guiding the agent towards the goal, irrespective of the initial start point.
 
In Figure~\ref{fig:metaworldsac}(a,b,c), we examine the outcomes of training three Meta World environments~\cite{metaworld} using the SAC algorithm. These results are compared against a sparse reward and an episode-specific dense reward. Notably, we encountered difficulties in implementing hindsight experience replay with these environments. The Figure~\ref{fig:metaworldsac}(a) represents the results of the open door task, (b) displays the results of the peg insertion task, and (c) showcases the peg assembly task. Across all these environments, we observed that SAC was unable to solve the tasks effectively when utilizing sparse rewards. For the open door and peg insertion task \name\ performs similar or better than the dense reward. The dense rewards have been finely tuned to each task and can require a lot of trial and error to finalize, while \name\ is a general reward that can be applied to any method. Regarding the peg assembly task, we discovered that SAC was unable to solve the task even with the dense reward during the training period. This is due to increased complexity of the task, which results in a noisier reward signal. This task is more difficult because there are a greater number of objectives to be accomplished. However, our method proved capable of swiftly solving the task despite these challenges.

\label{exp:tasks}


\subsubsection{One Expert Demonstration}
Next, we look at our method compared to baselines that require just 1 example. We compare against AWAC, AWAC + MCAC, SAC + RB and SAC + MCAC when just one expert demonstration is used. The results of this are shown in Figure~\ref{fig:all1}. We find that for the three meta world tasks our approach significantly outperforms the other approaches. AWAC and AWAC + MCAC work in the Peg Assembly and Door Open tasks, but they are not able to solve the problem as well as our approach is. For the RS-Door Open task, we find that our approach performs very similar to all other baselines. This is likely because the task is very easy and requires only a small amount of data. The RS-Lift task results look similar to the MetaWorld results where our method significantly outperforms the others. This task was more difficult than others due to the rotation of the block being different in each episode, but our method is able to handle it nonetheless. 



\subsubsection{100 Expert Demonstrations}
Finally, we look at our method compared to the same baselines when the baselines use 100 expert demonstrations. The results are shown in Figure~\ref{fig:all2}. Note that AWAC, which is the method that performs the best in these scenarios has to be pre-trained in addition to the online training. We find that although our method takes more online training time with just one demonstration, versus 100 demonstrations, our method performs equal to or better than the baselines in all of the tasks.
\section{Conclusion}
\label{sec:conclusion}
We present \name, a new approach that enables training reinforcement learning models using a single demonstration. Unlike other imitation learning methods, which typically rely on multiple demonstrations and access to detailed action information, \name\ can operate with limited information and single demonstration. In order to achieve this, we introduce a novel universal reward function and leverage a knowledge expansion technique that extends beyond initial start and goal states. This makes it highly suitable for learning tasks with minimal information across different environments. We show that \name\ is faster than standard reinforcement learning models, in cases where the rewards are sparse or dense, and showcase its ability to succeed where other approaches fall short. Additionally, we show that \name\ performs similar to or better than a variety of imitation learning methods when these methods use either one or one hundred expert demonstrations. In future research, we aim to explore the use of expansion for non-linear states and investigate the utilization of image-based state spaces rather than joint locations. 

\noindent\textbf{Acknowledgements:} This work was partially supported by DARPA SAIL-ON (W911NF2020009) program and NSF CAREER Award (\#2238769) to AS.


\clearpage


\printbibliography

\end{document}